\documentclass{article}

\usepackage{arxiv} 
\usepackage{natbib}
\usepackage[utf8]{inputenc} 
\usepackage[T1]{fontenc}    
\usepackage{hyperref}       
\usepackage{url}            
\usepackage{booktabs}       
\usepackage{amsfonts}       
\usepackage{nicefrac}       
\usepackage{microtype}      
\usepackage{xcolor}         
\usepackage[pdftex]{graphicx}
\usepackage{subcaption}
\usepackage{bbm} 
\newcommand{\comment}[1]{\ignorespaces}
\newcommand{\todo}[1]{}

\title{Realistic Handwritten Multi-Digit Writer (MDW) Number Recognition Challenges}

\author{%
  Kiri L.~Wagstaff\\
  Corvallis, OR 97330 \\
  \texttt{wkiri@wkiri.com} \hspace{0.2in} \url{https://www.wkiri.com/}\\
}

\begin{document}

\maketitle

\begin{abstract}
  Isolated digit classification has served as a motivating problem for
  decades of machine learning research.  In real settings, numbers
  often occur as multiple digits, all written by the same person.
  Examples include ZIP Codes, handwritten check amounts, and
  appointment times. 
  In this work, we leverage knowledge about the writers of NIST digit
  images to create more realistic benchmark multi-digit writer (MDW)
  data sets.  As expected, we find that classifiers may perform well
  on isolated digits yet do poorly on multi-digit number recognition.
  If we want to solve real number recognition problems, additional
  advances are needed.
  The MDW benchmarks come with task-specific performance metrics that
  go beyond typical error calculations to more closely align with
  real-world impact.  They also create opportunities to develop
  methods that can leverage task-specific knowledge to improve
  performance well beyond that of individual digit classification
  methods. 
\end{abstract}


\section{Introduction and related work}

The MNIST handwritten digit data set~\citep{lecun:mnist98}, derived
from the original data~\citep{wilkinson:nist92,grother:nist-sd19}
collected by the National Institute of Standards and Technology
(NIST), 
has been used extensively as a benchmark to compare classifiers.
Performance on the standard test set of 10,000 digits has reached
99.87\%~\citep{byerly:mnist21}, 
rendering it effectively a solved, and perhaps therefore less
interesting, problem.
Investigators subsequently developed several new data sets
(similarly formatted for convenience) that have different classes,
properties, and difficulty levels.  These include notMNIST
(letters A to J)~\citep{bulatov2011notmnist}, EMNIST (digits and
letters)~\citep{cohen:emnist17}, Fashion-MNIST (articles of
clothing)~\citep{xiao:fashion-mnist17}, Kuzushiji-MNIST (Japanese
hiragana)~\citep{clanuwat:kmnist18}, and more. 
These data sets all focus on individual digit (or item) classification.

However, recognizing isolated digits by randomly chosen writers is not
a very realistic task.
Multi-digit numbers are everywhere around us.
The Street View House Number (SVHN) data set~\citep{netzer:svhn11}
provides compelling real-world examples that were cropped from Google
Street View images.  It has inspired major advances in multi-digit number
recognition (e.g.,~\cite{goodfellow:svhn14}).  SVHN consists of printed
numbers, and there is a need for similar resources to inspire progress
on handwritten multi-digit numbers.

\begin{figure}
  \centering
  \begin{subfigure}{0.4\linewidth}
    \includegraphics[width=\linewidth]{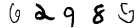}
    \caption{Each digit from a randomly chosen NIST writer}
    \label{fig:zip-rand}
  \end{subfigure}
  \hfill
  \begin{subfigure}{0.4\linewidth}
    \includegraphics[width=\linewidth]{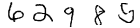}
    \caption{All digits from NIST writer 2445}
    \label{fig:zip-same}
  \end{subfigure}
  \caption{What is the ambiguous final digit?  (a) Random
    choices for the first four digits provide no clues.  (b) Knowing that
    the 9 in the middle position was created by the same NIST writer
    increases our confidence that the final digit is not a 9.  Indeed,
    it was labeled as a 5.}
  \label{fig:zip}
\end{figure}

To date, no standard benchmark data sets exist that contain
{\bf multi-digit handwritten numbers}, which has limited advances in this
arena. While it has been possible for
decades to re-mix the MNIST digits into random sequences, it was not
possible to ensure that a sequence 
came from a single source (as in SVHN),
because the information about which writer wrote each digit was not
preserved in the MNIST benchmark data set.  
Randomly generated digit sequences are not very realistic (compare
Figure~\ref{fig:zip-rand} to Figure~\ref{fig:zip-same}). 
Fortunately, later researchers invested painstaking effort to reconstruct
how the MNIST images were selected and processed, enabling each image
to be reassociated with its meta-data (including its writer) and published as the QMNIST data set~\citep{yadav:qmnist19}. 

This paper contributes three new MDW (multi-digit writer) benchmark
data sets inspired by real-world number recognition tasks.  We used
the writer meta-data in QMNIST to generate realistic digit sequences,
each created by a single writer.
The MDW data sets include U.S. ZIP Codes, check amounts, and times of day.
Each problem domain comes with natural constraints on valid numbers
and custom evaluation metrics tailored to the relevant use cases.
Because the sequences were assembled
from the 28x28 pixel digits in QMNIST, these test sets can be
immediately used by existing classifiers trained on MNIST data.
However, they also allow future handwritten number recognition
models to do more than simply classify each digit in isolation.
Future methods could leverage the knowledge that the same writer
generated each digit.  Ideally, this will lead to new advances that
improve digit disambiguation and overall number recognition accuracy.
\comment{
These tasks necessarily violate the standard assumption that data is
independent and identically distributed (i.i.d.).  We posit that
models that are able to leverage task-specific constraints and item
dependencies (e.g., the knowledge that all digits within a number were
written by the same person) can improve performance on these harder
problems.
}

A newly proposed benchmark should do more than present a new puzzle or
target to occupy time and effort.  Ideally, it can lead us to greater
understanding of the relevant problems as well as the strengths and
weaknesses of different solutions.  We propose going beyond simple
error metrics to incorporate realistic measurement of errors and their
impact~\citep{wagstaff:matters12}.
For example, can a ZIP Code classifier inadvertently exhibit
geographical bias?  We show how to investigate these and other
impact-related questions. 

Our data generation scripts can also be adapted to create and populate
other handwritten number recognition tasks (e.g., social security
numbers, phone numbers, dates).  Our overall goal is to
inspire innovation and advances that can improve performance on these
challenging multi-digit number recognition problems.

\section{Background: The writers of NIST (and MNIST)}


In the late 1980s, the U.S.~Census Bureau asked NIST
to assist in creating a data set
that could be used to assess different approaches to optical character
recognition~\citep{wilkinson:nist92}.  A successful system could be an
important labor-saving device for processing handwritten census forms.
Over several years, NIST collected handwriting samples from thousands of
Census Bureau field personnel located throughout the United States. In
addition, they collected samples from 500 high school students in
Maryland to use as a challenging test from a new population.

The published NIST Special Database 19 (SD-19) data set contains
3.9 million images of individual handwritten letters and digits that
were created by 3,650 writers. SD-19 is the result of 
several batches of data collection~\citep{grother:nist-sd19}.  Each
writer was asked to complete 
a Handwriting Sample Form (HSF) with entries for 130 digits, 
encompassing 13 versions of each digit from 0 to 9.
However, the SD-19 data set does not contain all 130 digits from every
writer, possibly due to incomplete forms or illegible entries.  On
average, there are 113 digits per writer.  The original digit images
are 128x128 pixels.
Each data collection effort is identified by a partition name (hsf\_0,
hsf\_1, ...).  The partition hsf\_4 contains the data collected
from high school students, while all other partitions contain Census
employee data.

An initial evaluation of 46 different classifiers from 26
organizations found that the high school student sample was
significantly more difficult than the Census employee training
set~\citep{wilkinson:nist92}. Not only were the writers from a
different demographic group, but different segmentation methods were used to
extract individual digits from each group's handwritten forms.
\cite{lecun:mnist98} addressed this by creating a Modified NIST
(MNIST) data set that included both Census and high school student
digits in the training and test sets.  They carefully segregated the writers so
no writer had digits in both the training and test sets.  The MNIST
training set contains 30,000 images from each group (60,000 total).
The public MNIST test set contains 5,000 images from each group
(10,000 total), with an additional 50,000 images reserved as an
internal test set.
%
LeCun et al.~also converted the 128x128 images into smaller 28x28
images for MNIST, likely due to computational constraints of the
time.


The QMNIST data set resurrected the per-digit meta-data by carefully
matching each MNIST digit image with its source NIST
image~\citep{yadav:qmnist19}.  They were even able to recreate the
hidden test set of 50,000 items. In total, QMNIST (402,953 images) is
much larger than the MNIST sample.  It consists of digits from 3,579
(vs. 1,074) writers.  The number of digits per writer ranges from 12 to
148, with up to 29 images of the same digit by a given writer.
For each digit image, QMNIST provides the writer id, class (label),
HSF (partition) id, and a global NIST index.  The writer and HSF
information could 
inform some very interesting error analysis and help determine whether some
writers have more difficult handwriting or which data collection
protocols led to data that was easier or more difficult to classify.
To date, these remain open questions.

To enable the creation of new benchmark test sets from the QMNIST
data, we first split the full set of NIST writers into training and test
writers.  For full compatibility with MNIST, we initialized
$\mathcal{W}_{tr}$ (training) and $\mathcal{W}_{te}$ (test) with the
539 and 535 (respectively) writers employed in the MNIST training and
test sets, then randomly split the remaining (previously unused) 2,505
writers between $\mathcal{W}_{tr}$ and $\mathcal{W}_{te}$.

\section{Multi-digit writer (MDW) number recognition benchmark test
  sets}
\label{sec:mdw}

Numbers in the real world rarely consist of a sequence of digits that
were each written by a different person. 
When scanning ZIP Codes or dollar amounts, we expect that all of the
digits in a number were created by the same writer. Knowledge about
which writer created each NIST digit enables us to construct novel,
realistic benchmark test sets.  Each benchmark also comes with
domain-specific performance metrics and constraints that could be
leveraged to improve number recognition performance.

Each MDW benchmark test set was created using the same methodology.
Code to generate these data sets (or inspire your own variations) is
available at \url{https://github.com/wkiri/MDW-handwritten/}.
We generated $m$ multi-digit numbers as follows.
Each number is represented by a sequence of individual digit images 
$\{ I_1, I_2, \ldots, I_k\}$, where $k$ is dictated by the test
domain.
We first randomly chose a $k$-digit number $N = d_1 d_2 \ldots 
d_k$.  We also imposed domain-specific constraints to ensure that the
number $N$ is valid (e.g., valid U.S. ZIP Codes are a subset of all
possible five-digit numbers). 
Next, we randomly chose a test writer $w \in \mathcal{W}_{te}$, then
selected an image 
$I_i$ for each digit $d_i$ by randomly selecting from $D_{w,d_i}$, the
set of all images of digit $d_i$ by writer $w$.
We store a string representation of the multi-digit number, the writer
id, and the sequence of digit image identifiers in a .csv file for 
ease of use.

\begin{table}
  \caption{Multi-digit writer (MDW) number recognition benchmarks}
  \label{tab:data}
  \centering
  \setlength\tabcolsep{4pt} 
  \begin{tabular}{llll}
    \toprule
    \textbf{Benchmark} & \textbf{Items} & \textbf{Digits per item} & \textbf{Range} \\
    \midrule
    MDW-ZIP-Codes & 10,000 & 5 & 00601 to 99926 \\
    MDW-Check-Amounts & 10,000 & 3 to 7 & \$0.05 to \$76,805.81 \\
    MDW-Clock-Times & 10,000 & 3 to 4 & 0:00 to 23:58 \\
    \bottomrule
  \end{tabular}
\end{table}

Table~\ref{tab:data} summarizes the MDW number
recognition benchmark data sets that we have constructed.  They can be
downloaded from
\url{https://www.kaggle.com/datasets/kirilwagstaff/multi-digit-writer-mdw-number-recognition}.
Any classifier trained on MNIST 28x28 images can be applied immediately
to these benchmark data sets.  One needs only to load the QMNIST data set with extended
labels, then use the MDW benchmark test set files to obtain the
sequence of images associated with each multi-digit number (see
Section~\ref{sec:use} for how to use our scripts as examples).  A
starting approach is to classify each digit
individually and independently. We evaluated some standard machine
learning classifiers on these benchmarks for illustration.
However, it is our hope that these results will inspire new
classifiers that leverage known task-specific constraints about valid
numbers and the context provided by other digits within the number. 

\subsection{MDW-ZIP-Codes}
\label{sec:zip}

We expect that a handwritten U.S. ZIP Code is a sequence of five digits,
all written by the same person. (Globally, postal codes vary
in length and may also include letters.)
For this application, performance on all $10^5$ possible combinations
of five digits would not be a useful metric, because only 37,988 are
in use, fewer than half of the possibilities~\citep{zipcodes25}.
We constructed a benchmark test composed of 10,000 samples of valid
U.S. ZIP Codes.

\begin{figure}[b]
\centering
\begin{subfigure}{0.67\linewidth}{
    \begin{minipage}[t]{\textwidth}
      \resizebox{\linewidth}{!}{ 
        \setlength\tabcolsep{4pt} 
        \begin{tabular}{lllllll}
          \toprule
          Numeral & Writer & NIST id 0 & NIST id 1 & NIST id 2 & NIST id 3 &
          NIST id 4 \\
          \midrule
          27892 & 3688 & 353625 & 353544 & 353630 & 353613 & 353625 \\
          \bottomrule
        \end{tabular}
      }
    \end{minipage}
  }
\end{subfigure}
\hfill
\begin{subfigure}{0.3\linewidth}
  \includegraphics[width=\linewidth]{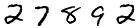}
\end{subfigure}
\caption{Example 5-digit item (by writer 3688) from MDW-ZIP-Codes and
  its visualization.}  
\label{fig:ex-zip}
\end{figure}

\paragraph{ZIP Code generation.} To create each test item, we randomly
selected a ZIP Code $z$ from the list of valid entries and randomly chose
a writer $w$ from $\mathcal{W}_{te}$. For each digit $d$ in $z$, we
randomly chose an image of digit $d$ from $D_{w,d}$.  Each item in
the data set consists of the ZIP Code $z$ (as a string), NIST writer id
$w$, and five NIST image ids.
For example, one entry in the data set is for ZIP Code 27892,
with all digits written by writer 3688, and the remaining columns
specify the individual NIST image ids for each digit (see
Figure~\ref{fig:ex-zip}).  

\begin{figure}
  \centering
  \begin{subfigure}{0.5\linewidth}
    \includegraphics[height=1.7in]{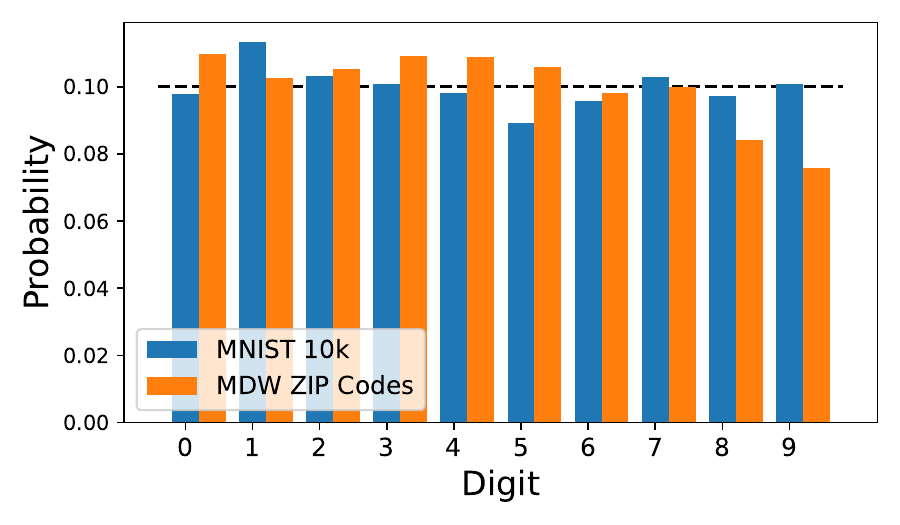}
    \caption{Digit distribution for MNIST test set and MDW-ZIP-Codes} 
    \label{fig:zip-dist}
  \end{subfigure}
  \hfill
  \begin{subfigure}{0.45\linewidth}
    \includegraphics[height=1.7in]{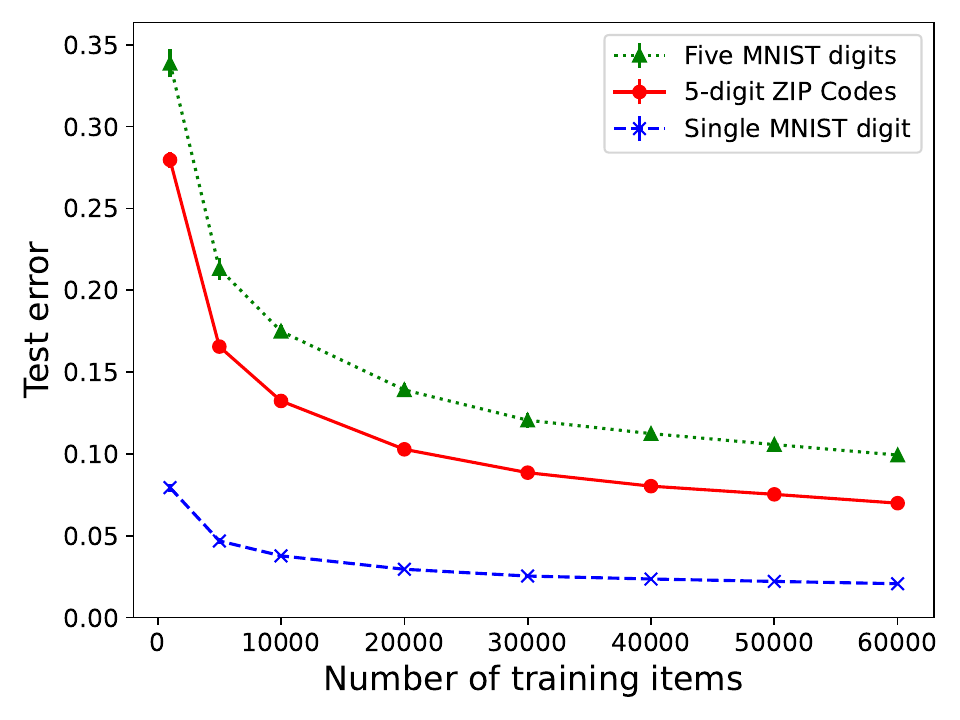}
    \caption{SVM learning curves for recognizing single digits (easier)
      and 5-digit ZIP Codes (harder)}
    \label{fig:zip-learn}
  \end{subfigure}
  \caption{Digit distribution and learning curves for
    the MNIST test set versus U.S. ZIP Codes.}
  \label{fig:zipcode}
\end{figure}

The resulting collection of 50,000 digits in 10,000 ZIP Code samples
has a digit distribution that differs from that of the MNIST test set.
Figure~\ref{fig:zip-dist} shows that both distributions exhibit class
imbalance, but the class proportions are different.  The most common
digits in the MNIST test set are 1, 2, 7, and 9, while the most common
ZIP Code digits are 0, 3, 4, and 5.  Digits 8 and 9 are particularly 
under-represented in ZIP Codes.

\paragraph{ZIP Code performance metrics.}  In this application, successful ZIP
Code recognition requires that every individual digit is correct.  If
not, the item being mailed could be routed to an incorrect
destination.  Given classifier $\mathcal{C}$ that makes five-digit predictions
$\mathcal{C}(z)$ for ZIP Code image sequence $z$, we calculate the
classifier's error on ZIP Code test set $T$ with five-digit labels $y_z$ for
$z \in T$ as

\[ Err_{strict}(\mathcal{C}, T) = \frac{1}{|T|} | \{z \in T \mid
\mathcal{C}(z) \neq y_z\} | \] 

Recognizing five-digit numbers is, unsurprisingly, harder than
recognizing individual digits (see Figure~\ref{fig:zip-learn}).
These results were obtained over 10 trials (error bars show standard
deviation) in which we sampled successively larger MNIST training sets to
train a support vector machine (SVM) and evaluated it
on the fixed test sets. The ``Single MNIST digit'' error is the
fraction of  MNIST 10k test digits misclassified when classified
individually.  We calculate the expected $Err_{strict}$ for five
digits, given an MNIST single-digit error rate of $e$, as $1 -
(1-e)^5$. This is shown as the ``Five MNIST digits'' curve; as
expected, it is much worse than single-digit performance. 

The ``5-digit ZIP Codes'' curve is the observed error rate
$Err_{strict}$ for the same SVM classifier on the MDW-ZIP-Codes test
set.  It is lower (better) than the expected error $1 - (1-e)^5$.  The expected
error is only a good estimate of general 5-digit performance if (1)
the new test set is drawn from the same distribution as the MNIST test
set and (2) the single-digit error rate $e$ is consistent across all
digit classes.
Differences in class distribution (Figure~\ref{fig:zip-dist}) and
per-class performance likely contribute to this result.

Not all errors have the same properties. 
A prediction $\mathcal{C}(z)$ that is an invalid ZIP Code can be
immediately flagged as an error, while mistakenly recognizing $z$ as a
different, but valid, ZIP Code is more subtle.  We calculate both
kinds of error with respect to $\mathcal{V}_{z}$, the set of valid
ZIP Codes, as

\[ Err_{invalid}(\mathcal{C}, T) = \frac{1}{|T|} | \{z \in T \mid
\mathcal{C}(z) \neq y_z, \mathcal{C}(z) \notin \mathcal{V}_{z}\} | \] 
\[ Err_{valid}(\mathcal{C}, T) = \frac{1}{|T|} | \{z \in T \mid
\mathcal{C}(z) \neq y_z, \mathcal{C}(z) \in \mathcal{V}_{z}\} | \]

where $Err_{strict}(\mathcal{C}, T) = Err_{invalid}(\mathcal{C}, T) +
Err_{valid}(\mathcal{C}, T)$.


\begin{table}
  \centering
  \caption{Error metrics for ZIP Code recognition
    (lower is better). The first column shows error rate on the
    standard MNIST single-digit test set ($n$=10,000), while the last
    three columns show error on the MDW-ZIP-Codes data set ($n$=10,000).
    Results for random forests are 
    averages over 10 values of the \texttt{random\_state} parameter,
    with the standard deviation shown in parentheses.
    The best result for each column is in bold.}
  \label{tab:res-zip}
  \small
  \begin{tabular}{lllll}
    \toprule
    MNIST single-digit error & Classifier & $Err_{strict}$ & $Err_{invalid}$ & $Err_{valid}$ \\
    \midrule
    0.0327 (0.0015) & Random forest (50 trees)  & 0.1132 (0.0017) &
    0.0559 (0.0015) & 0.0574 (0.0010) \\
    0.0307 (0.0006) & Random forest (100 trees) & 0.1029 (0.0015) &
    0.0500 (0.0010) & 0.0529 (0.0011) \\
    0.0289 (0.0007) & Random forest (200 trees) & 0.1013 (0.0010) &
    0.0500 (0.0010) & 0.0513 (0.0007) \\ 
    0.0207 & SVM (RBF kernel, $\gamma=$scale) & 0.0700 & 0.0332 & 0.0368 \\
    0.0140 & CNN (3 layers, 10 epochs) & 0.0436 & 0.0197 & 0.0239 \\
    0.0127 & CNN (5 layers, 20 epochs) & 0.0415 & 0.0202 & 0.0213 \\
    {\bf 0.0083} & VGG-like CNN (7 layers, 20 epochs) & {\bf 0.0299} & {\bf 0.0148} & {\bf 0.0151} \\
    \bottomrule
\end{tabular}
\end{table}

To illustrate how to use this benchmark, we evaluated several
single-digit classifiers by applying them to each digit and
concatenating their predictions to generate a 5-digit number (see
Table~\ref{tab:res-zip}).  Each model was trained on the full MNIST
training set (60,000 items).  The models include a random forest with
50, 100, or 200 trees, a support vector machine with a radial basis
function (RBF) kernel,  
convolutional neural networks (CNN) with 3 or 5 layers ([Conv2d+ReLU]
x 3 + Linear or [Conv2d+ReLU] x 5 + Linear) as well as a CNN inspired
by VGG ([Conv2d+ReLU+MaxPool2d] x 2 + Linear).  The 28x28 images are
too small to be used with the full-size VGG architecture.  Each CNN
was trained for 10 or 20 epochs, the point at which held-out
performance stopped improving.

The VGG-like CNN achieved the lowest error rate for single-digit
classification (0.0083). It also demonstrated the best multi-digit
recognition in all three metrics, but the error rate was much higher
(0.0299).  For all classifiers, there were slightly more valid errors than
invalid errors.  To aid deployment decisions, both of these error
rates should be measured, since invalid errors can be immediately
detected and receive further processing.
Based on these results, we anticipate that
even the current best single-digit classifiers would likewise exhibit
room for improvement for the ZIP Code benchmark test set.

\paragraph{ZIP Code geographical bias.} Some digits, and therefore
some ZIP Codes, are more difficult than others to classify.  Could a
classifier's weaknesses on certain digits 
inadvertently introduce a geographical bias into an operational postal
service deployment?  Because ZIP Codes are associated with physical
locations, we can check whether a classifier exhibits such a bias.

\begin{figure}
  \centering
  \begin{subfigure}{0.49\linewidth}
    \includegraphics[height=1.7in]{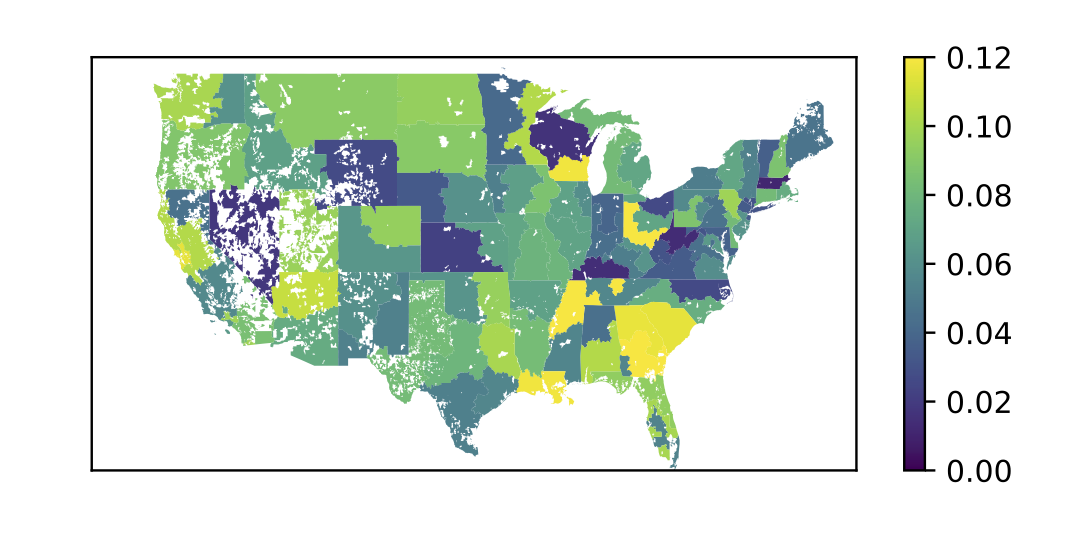}
    \caption{SVM (RBF kernel).
      Error varies from 0.01 (sector 01XXX, Massachusetts) to 0.13
      (sector 00XXX, Puerto Rico).}
  \end{subfigure}
  \hfill
  \begin{subfigure}{0.49\linewidth}
    \includegraphics[height=1.7in]{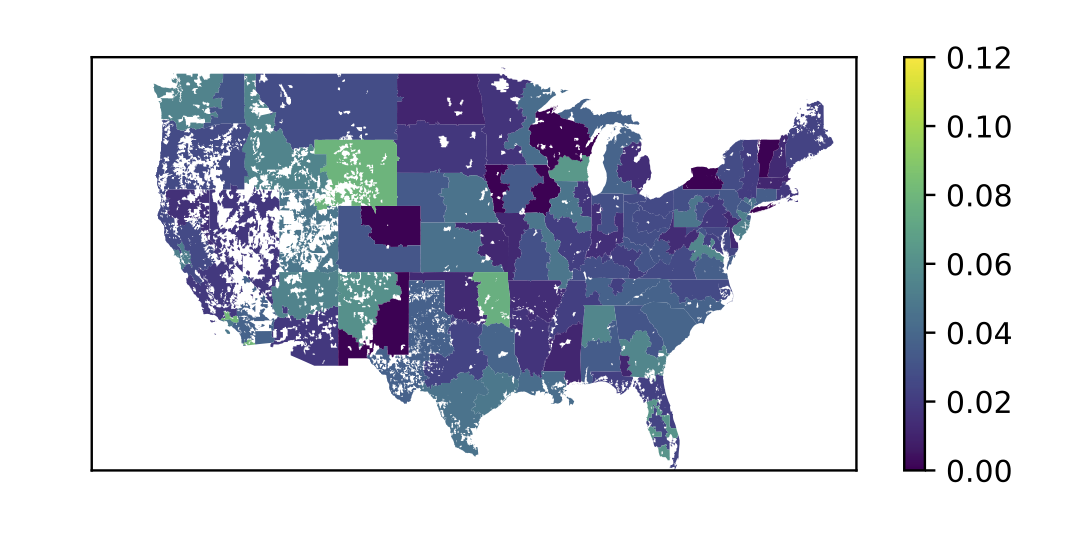}
    \caption{VGG-like CNN (20 epochs).
      Error varies from 0.00 (sector 05XXX, Vermont)
      to 0.08 (sector 91XXX, Los Angeles, CA).}
  \end{subfigure}
  \begin{subfigure}{0.52\linewidth}
    \includegraphics[height=1.7in]{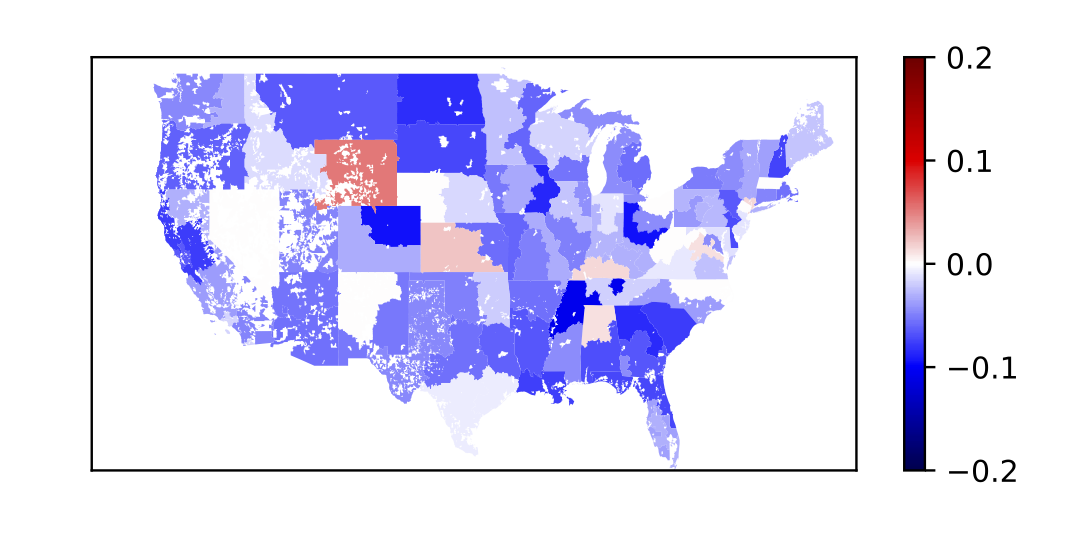}
    \caption{Difference in sector error rate:
      red (positive) values show sectors where the
      VGG-like CNN had a higher error rate than the SVM.}
  \end{subfigure}
  \caption{Geographical distribution of U.S. ZIP Code recognition
    error for two classifiers, per ZIP Code sector.  The
    VGG-like CNN had the lowest nation-wide error rate (0.03 vs.~0.07),
    but it showed a much higher error rate than the SVM in specific sectors.}
  \label{fig:spatial-zip}
\end{figure}

U.S. ZIP Codes are organized hierarchically.  We used the first two
digits of each ZIP Code to define a geographical sector that
spans many ZIP Codes (e.g.,~00XXX, 01XXX, 02XXX, \ldots).
We computed a sector-level error rate based on all test items that
fell within each sector. 
Figure~\ref{fig:spatial-zip} shows the map of ZIP Code error rates
($Err_{strict}$) per sector for the SVM and VGG-like CNN after
training on the MNIST training set.  White areas do not belong to any
U.S.~ZIP Code.
We find high variability across sectors, and the classifiers differ
in which sectors they find most challenging.

The choice of which model to use operationally could have significant
impacts for residents of different areas.
Figure~\ref{fig:spatial-zip}(c) shows the difference in error rates.
While the VGG-like CNN has the best performance
when averaged across the nation, its error rate in 
Colorado and Wyoming (sector 82XXX) is 5\% higher than that of the
SVM in the same sector.  Conversely, in Puerto Rico (sector 00XXX),
the SVM is a much worse choice (11\% higher error rate than the
VGG-like CNN).  If faced with these options, the 
U.S. Postal Service might choose to deploy different models in different
locations to reduce geographical bias in recognition errors
that unevenly impact different locations.
To our knowledge, this is the first time anyone has checked for
geographical bias in digit (or ZIP Code) classifiers.

\subsection{MDW-Check-Amounts}

One of the original motivating applications for the creation of the
NIST digit recognition data set was reading the transaction amount
from a handwritten check~\citep{lecun:mnist98}.  Checks contain two
representations of the amount: a legal amount which is written
out in words, and a courtesy amount which is written with
numerals.  We focus on the task of accurately recognizing the courtesy
amount. 
In this setting, the number of digits is variable, and they are all
written by the same author.  We constructed a benchmark test set
composed of 10,000 artificial check amounts in U.S. dollars.

\paragraph{Check amount generation.}
We used the Newcomb-Benford law~\citep{newcomb:law1881,benford:law38} as
the generating distribution.  This law captures the curious empirical
observation that digits in naturally occurring numbers are not
uniformly distributed (in fact, deviation from this law is
used to detect batches of potentially fraudulent
transactions~\citep{nigrini:fraud12}). 
For this distribution, the probability $P_n$ of the first
digit of a number being $n$ is $\log \left(1 + \frac{1}{n} \right)$.

\begin{figure}[b]
\centering
  \small
  \setlength\tabcolsep{4pt} 
\begin{tabular}{rllllllll}
  \toprule
  Numeral & Writer & NIST id 0 & NIST id 1 & NIST id 2 & NIST id 3 &
  NIST id 4 & NIST id 5 & NIST id 6\\
  \midrule
  7680581 & 3797 & 366515 & 366524 & 366565 & 366508 & 366528 & 366577
  & 366576 \\
  752 & 3598 & -1 & -1 & -1 & -1 & 342695 & 342627 & 342668 \\
  \bottomrule
\end{tabular}
  \includegraphics[height=0.3in]{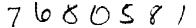}
  \hspace{0.75in}
  \includegraphics[height=0.3in]{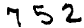}
  \caption{Two check amount items from MDW-Check-Amounts (\$76,805.81
    and \$7.52) and their visualizations.  -1 indicates no digit,
    allowing for variable length numbers.} 
  \label{fig:ex-check}
\end{figure}

To create each test item, we used the Python \texttt{randalyze}
library~\citep{ross:randalyze23} to generate amounts between \$0.00
and \$99999.99 (1 to 5 digits before and 2 digits after the decimal
place).
%
To store the amounts and their NIST image ids in a file, we omit the
decimal and store the amount in cents (3 to 7 digits). Since the
number of digits is variable, we pad the list of NIST ids by
prepending a -1 for each unused digit.  For example, the two entries
in the benchmark file for \$76,805.81 by writer 3797 and \$7.52 by
writer 3598 are shown in Figure~\ref{fig:ex-check}. 

\paragraph{Check amount performance metrics.} As in the case of ZIP
Code recognition, we calculate $Err_{strict}$ and partition it into
invalid and valid errors.  In this domain, the only invalid errors are
predictions that have a leading zero and more than three digits.  An
amount of \$0.50 is valid, but an amount of \$047.22 is not.

In addition, for check amount recognition, the impact of a digit
classification error varies depending on its position in the number.
An error in the number of cents may not matter as much as an error in
the thousands of dollars.
We convert each predicted digit string back into cents, then
divide by 100 to obtain dollars.  We calculate the total (absolute)
error over the data set $Err_{total}$ and the largest single
(absolute) error $Err_{max}$.  The sign of each error also matters.
If a check amount is recognized as larger than its true value, the
person who wrote the check could experience an overdraft on their
banking account.  If the amount is recognized as smaller than its true
value, they could incur late fees or other underpayment penalties.
We assess whether a classifier tends to over- or under-predict amounts
by calculating the average per-check (signed)  $Err_{avg}$.

\begin{table}
  \centering
  \caption{Error and cost metrics for check amount recognition
    (lower is better, $n$=10,000). Results for random forests are the
    average over 10 values for the \texttt{random\_state} parameter,
    with the standard error in parentheses.
    The best result for each metric is in bold.}
  \label{tab:res-check}
  \small
  \begin{tabular}{llll}
    \toprule
    Classifier & $Err_{strict}$ & $Err_{invalid}$ & $Err_{valid}$ \\
    \midrule
    Random forest (50 trees) & 0.1179 (0.0019) & 0.0018 (0.0002) &
    0.1161 (0.0020) \\
    Random forest (100 trees) & 0.1101 (0.0018) & 0.0016 (0.0002) &
    0.1084 (0.0017) \\
    Random forest (200 trees) & 0.1071 (0.0005) & 0.0017 (0.0001) &
    0.1054 (0.0005) \\
    SVM (RBF kernel, $\gamma=$scale) & 0.0786 & 0.0012 & 0.0774 \\
    CNN (3 layers, 10 epochs) & 0.0499 & 0.0010 & 0.0489 \\
    CNN (5 layers, 20 epochs) & 0.0460 & {\bf 0.0004} & 0.0456 \\
    VGG-like CNN (7 layers, 20 epochs) & {\bf 0.0332} & 0.0006 &
    {\bf 0.0326} \\
    \midrule
    Classifier & $Err_{total}$ & $Err_{avg}$ & $Err_{max}$ \\
    \midrule
    Random forest (50 trees) & \$626,131.72 (\$55,655) & \$37.11
    (\$4.86) & \$61,012.00 (\$7,358.79) \\
    Random forest (100 trees) & \$571,741.93 (\$36,605) & \$34.81
    (\$4.81) & \$62,012.00 (\$7,867.84) \\
    Random forest (200 trees) & \$549,518.51 (\$36,272.81) & \$34.19
    (\$3.59) & \$64,012.00 (\$8,410.61) \\
    SVM (RBF kernel, $\gamma=$scale) & \$434,309.73 & \$27.67 & \$70,000.00 \\
    CNN (3 layers, 10 epochs) & \$465,379.34 & \$40.84 & \$60,000.00 \\
    CNN (5 layers, 20 epochs) & {\bf \$167,969.21} & {\bf \$9.31}  & {\bf \$30,000.00} \\
    VGG-like CNN (7 layers, 20 epochs) & \$200,693.75 & \$15.88 & \$60,000.00 \\
    \bottomrule
\end{tabular}
\end{table}

We demonstrated use of this benchmark with the same set of
classifiers (Table~\ref{tab:res-check}).
The VGG-like CNN achieved the lowest error for the strict and valid
error metrics.  Invalid errors, with an amount containing an illegal
leading zero, were rare (as expected) and also would be easily
flagged.  The 5-layer CNN achieved the lowest error rate for this
metric. 
In terms of cost metrics, we found that the 5-layer CNN out-performed
the VGG-like CNN, achieving the lowest total, average, and maximum cost.
This kind of evaluation can
inform decisions about which model to employ, depending on the
relative importance of each kind of cost.
All models showed a bias towards over-estimating check amounts, rather
than under-estimating them; all $Err_{avg}$ values were positive.
None of these models make use of our prior knowledge about the
Newcomb-Benford law that governs digit frequencies.  We posit that
models taking this knowledge into account could greatly reduce the
most costly errors. 


\subsection{MDW-Clock-Times}

A final common case in which handwritten numbers are used is to write
down the time of an appointment or important event.  We focus on
24-hour clock time, which ranges from 0:00 (midnight) to 23:59 (the last
minute of the day).  We assume that times earlier than 10:00 omit the
leading digit, so the number of digits ranges from 3 to 4. We
constructed a benchmark test set composed of 10,000 artificially
generated times. 

\begin{figure}[b]
\centering
\begin{subfigure}{0.67\linewidth}{
    \begin{minipage}[t]{\textwidth}
      \resizebox{\linewidth}{!}{ 
        \begin{tabular}{lllllll}
          \toprule
          Numeral & Writer & NIST id 0 & NIST id 1 & NIST id 2 & NIST id 3 \\
          \midrule
          612 & 824 & -1 & 88310 & 88261 & 88271 \\
          \bottomrule
        \end{tabular}
      }
    \end{minipage}
  }
\end{subfigure}
\hfill
\begin{subfigure}{0.2\linewidth}
  \includegraphics[width=\linewidth]{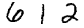}
\end{subfigure}
\caption{Example 3-digit item from MDW-Clock-Times representing 6:12.}
\label{fig:ex-clock}
\end{figure}

\paragraph{Time generation.}  To ensure the generation of valid clock times,
we randomly and uniformly sampled each time $t$ from 0 to 1,439
minutes and then converted $t$ to hours and minutes to obtain the time
as H:MM or HH:MM, where H or HH is the number of hours and MM is the
two-digit number of minutes past the hour.  As with check amounts, we
omit the punctuation (colon) with the understanding that a 3-digit
number HMM represents H:MM, and a 4-digit number HHMM represents
HH:MM.  For example, the entry for 6:12 by writer 824 is shown in
Figure~\ref{fig:ex-clock}. 

\paragraph{Time performance metrics.}
In this domain, invalid errors occur for predicted times that have an
hour greater than 23 or minutes greater than 59.
We calculate strict error, invalid error, and valid error rates.
Digit position is meaningful in this domain as well, so we also
calculate the magnitude and sign of numeric errors (in
minutes) with $Err_{total}$, $Err_{avg}$, and $Err_{max}$.

\begin{table}
  \centering
  \caption{Error and cost metrics for clock time recognition
    (lower is better, $n$=10,000). Results for random forests are the
    average over 10 values for the \texttt{random\_state} parameter,
    with the standard error in parentheses.
    The best result for each metric is in bold.}
  \label{tab:res-clock}
  \small
  \begin{tabular}{llll}
    \toprule
    Classifier & $Err_{strict}$ & $Err_{invalid}$ & $Err_{valid}$ \\
    \midrule
    Random forest (50 trees)  & 0.0821 (0.0019) & 0.0207 (0.0010) &
    0.0614 (0.0014) \\
    Random forest (100 trees) & 0.0774 (0.0007) & 0.0195 (0.0006) &
    0.0579 (0.0006) \\
    Random forest (200 trees) & 0.0753 (0.0011) & 0.0192 (0.0009) &
    0.0561 (0.0007) \\
    SVM (RBF kernel, $\gamma=$scale) & 0.0515 & 0.0131 &
    0.0384 \\
    CNN (3 layers, 10 epochs) & 0.0309 & 0.0109 & 0.0200 \\
    CNN (5 layers, 20 epochs) & 0.0309 & 0.0088 & 0.0221 \\
    VGG-like CNN (7 layers, 20 epochs) & {\bf 0.0229} & {\bf 0.0080} &
    {\bf 0.0149} \\
    \midrule
    Classifier & $Err_{total}$ (minutes) & $Err_{avg}$ (mins) & $Err_{max}$
    (mins) \\
    \midrule
    Random forest (50 trees)  & 65,149.3 (4,003.1) & -0.907 (0.3) &
    1204.1 (8.4) \\
    Random forest (100 trees) & 62,191.9 (1,941.6) & -0.642 (0.2) &
    1202.0 (6.3) \\
    Random forest (200 trees) & 60,340.7 (1,520.4) & -0.635 (0.1) &
    1200.0 (0.0) \\
    SVM (RBF kernel, $\gamma=$scale) & 36,549.0 & -0.600 & 1200.0 \\
    CNN (3 layers, 10 epochs) & 19,941.0 & -0.300 & 1200.0 \\
    CNN (5 layers, 20 epochs) & 21,045.0 & {\bf -0.260} & {\bf 600.0} \\
    VGG-like CNN (7 layers, 20 epochs) & {\bf 16,688.0} & 0.41 & 660.0 \\
    \bottomrule
\end{tabular}
\end{table}

We again found that the VGG-like SVM classifier obtained the lowest
overall error 
(Table~\ref{tab:res-clock}). Despite the large potential space of
invalid times (anything larger than 23:59 or a time with minutes
greater than 59), prediction errors were
again more likely to be valid but mistaken times.
Most of the classifiers had a negative average error, indicating that
time was more likely to be under-estimated than over-estimated.  The
VGG-like CNN had the opposite tendency to over-estimate times.  The
5-layer CNN achieved the smallest average and maximum errors, but the
VGG-like CNN had the lowest total error across the test set.
Once again, incorporating knowledge about valid times and the
impact of positional errors into the process could help reduce the
largest errors.  This benchmark, and its associated metrics, can help
measure progress towards that goal.


\section{How to use the MDW benchmark test sets and code}
\label{sec:use}

The MDW multi-digit number recognition benchmark data sets do not contain
the individual digit image data, since it is already publicly
available.  Instead, they provide the image indices needed to perform
testing on the benchmark.
The MDW data sets can be downloaded from
\url{https://www.kaggle.com/datasets/kirilwagstaff/multi-digit-writer-mdw-number-recognition}.
The code needed to generate the data sets and scripts to make and
evaluate predictions, to enable comparison with other results on the
same data set, are available at
\url{https://github.com/wkiri/MDW-handwritten/}.  

\subsection{Replication of results}
\label{sec:replic}

The three MDW benchmark data sets were created as follows:

{\small
\begin{verbatim}
% export DATA=data-multidigit
% python3 create_MDW_data.py -d zip_code -n 10000 -s 0 -f $DATA/test_MDW_zip_code.csv
% python3 create_MDW_data.py -d check_amount -n 10000 -s 0 -f $DATA/test_MDW_check_amount.csv
% python3 create_MDW_data.py -d clock_time -n 10000 -s 0 -f $DATA/test_MDW_clock_time.csv
\end{verbatim}
}

Predictions were generated (e.g., using the RBF SVM classifier) with:

{\small
\begin{verbatim}
% python3 predict_MDW_data.py -c <classifier> <MDW_data_file> <preds_file>
% python3 predict_MDW_data.py -c SVM $DATA/test_MDW_zip_code.csv preds-zip-codes.csv
% python3 predict_MDW_data.py -c SVM $DATA/test_MDW_check_amount.csv preds-check-amounts.csv
% python3 predict_MDW_data.py -c SVM $DATA/test_MDW_clock_time.csv preds-clock-times.csv
\end{verbatim}
}

Results in Tables~\ref{tab:res-zip}, \ref{tab:res-check}, and
\ref{tab:res-clock} were obtained with:

{\small
\begin{verbatim}
% python3 eval_MDW_data.py -d <domain> <MDW_data_file> <preds_file>
% python3 eval_MDW_data.py -d zip_code $DATA/test_MDW_zip_code.csv preds-zip-codes.csv
% python3 eval_MDW_data.py -d check_amount $DATA/test_MDW_check_amount.csv preds-check-amounts.csv
% python3 eval_MDW_data.py -d clock_time $DATA/test_MDW_clock_time.csv preds-clock-times.csv
\end{verbatim}
}

The SVM learning curve in Figure~\ref{fig:zip-learn} was obtained with:

{\small
\begin{verbatim}
% python3 exp_learning_curve.py
\end{verbatim}
}

The geographical bias plots in Figure~\ref{fig:spatial-zip} leverage
a GeoJSON U.S. ZIP Code file~\citep{goodall:zcta5-json} and are
generated by running this experiment:

{\small
\begin{verbatim}
% python3 exp_geographical_bias.py $DATA/test_MDW_zip_code.csv preds-zip-codes.csv
\end{verbatim}
}

The difference plot in Figure~\ref{fig:spatial-zip}(c) can be
generated by providing two ZIP Code prediction files to this script:

{\small
\begin{verbatim}
% python3 diff_geographical_bias.py $DATA/test_MDW_zip_code.csv \
          preds-zip-codes-1.csv preds-zip-codes-2.csv
\end{verbatim}
}

All experiments were run on a MacBook Air with 8 GB of RAM. The data
generation scripts took about 2.5 hours to run for each domain.
Generating predictions for each benchmark test set took 4 minutes.
Evaluating the predictions took 3 seconds.  The learning curve
experiment used 8 training set sizes and 10 trials required training 80
models and evaluating each one, consuming roughly 5 hours total.

\subsection{Extensions}

The MDW benchmark generation script can be used to create additional
data sets with new samples from each domain.
Files are included that contain the NIST
writer ids in $\mathcal{W}_{tr}$ (training) and $\mathcal{W}_{te}$
(test). The \texttt{create\_MDW\_data.py} script takes a writer list
as input.  For example, specifying \texttt{writers-all-training.csv}
enables the generation of multi-digit training examples while avoiding
data leakage by ensuring that no test writers are used.
For reference, the package also includes files with the subset of
writer ids used for the MNIST training and test sets.

Further, these data generation scripts can be adapted to create
and populate other handwritten number recognition tasks (e.g., social
security numbers, phone numbers, dates). 

\comment{
Predictions will be saved into a .csv file (preds\_file) with one
recognized number (prediction) per line, in the same order as items in
the MDW data file.

You can also make use of the MDW benchmarks independently of these
scripts. First, load the QMNIST data
set~\citep{yadav:qmnist19}.  This data set is available through the
\texttt{torchvision} Python library.  Specify that you want the
\texttt{"nist"} data set to get the entire set of 402,953 labeled NIST
digit images. If you would like to be able to access the full
meta-data (including writer, partition, and other information),
specify \texttt{compat=False} to obtain the extended labels.

{\small
\begin{verbatim}
>>> import torchvision.datasets as datasets
>>> qmnist_data = datasets.QMNIST(root="./data", what="nist",
                                  compat=False, download=True)
>>> all_images, all_labels = zip(*qmnist_data)
\end{verbatim}
}

Each of the multi-digit benchmark test sets specifies which NIST
images comprise each number.
The images can be directly obtained using the NIST ids. For example,
the digit images for 6:12 by writer 825 (Figure~\ref{fig:ex-clock})
are obtained by:

{\small
\begin{verbatim}
>>> clock_digit_1 = all_images[88310]
>>> clock_digit_2 = all_images[88261]
>>> clock_digit_3 = all_images[88271]
\end{verbatim}
}

Any ids specified as $-1$ (such as NIST id 0 in this example) are
placeholders and should be skipped.  They are used in MDW benchmarks
with variable numbers of digits.
}

\section{Limitations}
\label{sec:limit}

Although the goal of this work is to create (more) realistic
multi-digit benchmarks, it only encompasses the number recognition
step.  The initial step of segmenting the number into digits is
bypassed.
Because the multi-digit numbers were created by concatenating
individual digits, they do not capture the dynamics of numbers written
sequentially at the same time.
In addition, the MDW benchmarks are limited to the contents of the
original NIST 
data set, so they only encompass handwriting for a collection of
Census Bureau employees and Maryland high school students in the
1990s.  The sample may not be diverse enough to provide a reliable
estimate of number recognition capability for numbers written by the
broader population.

\section{Conclusions and Future Directions}
\label{sec:conc}

The MDW benchmark data sets consist of multi-digit handwritten numbers
that occur in three natural domains: ZIP Codes, check amounts, and
clock times.  They go beyond isolated digit recognition to allow us to
assess the impact of successful (or unsuccessful) number recognition
for real tasks.  Thanks to NIST, the thousands of people who filled
Handwriting Sample Forms, \cite{lecun:mnist98},
and~\cite{yadav:qmnist19}, we are able to 
assemble numbers in which all digits were written by the same person.

This work aims to inspire advances in multi-digit number recognition
along with careful assessment of potential disparate impacts, such as
geographical bias in ZIP Code recognizers.  We found that different
models trained on the same data develop strengths and weaknesses that
manifest as large changes in which geographic regions experience large
or small error rates.
We look forward to innovations that can address these and related issues.
      
\section*{Acknowledgments}
  We are grateful to the hard work invested by NIST and the thousands
  of people who filled out Handwriting Sample Forms to create a data
  set with lasting influence and fascinating variety.
  We thank Zeke Wander for fruitful discussions about this data set,
  pointers to useful resources to enable geographical ZIP Code plots,
  and suggestions for improvement of the paper. 

\newpage
\bibliography{refs}

\begin{thebibliography}{18}
\providecommand{\natexlab}[1]{#1}
\providecommand{\url}[1]{\texttt{#1}}
\expandafter\ifx\csname urlstyle\endcsname\relax
  \providecommand{\doi}[1]{doi: #1}\else
  \providecommand{\doi}{doi: \begingroup \urlstyle{rm}\Url}\fi

\bibitem[Benford(1938)]{benford:law38}
Frank Benford.
\newblock The law of anomalous numbers.
\newblock In \emph{Proceedings of the American Philosophical Society},
  volume~78, pages 551--572, 1938.

\bibitem[Bulatov(2011)]{bulatov2011notmnist}
Yaroslav Bulatov.
\newblock {NotMNIST} dataset, 2011.
\newblock URL
  \url{http://yaroslavvb.blogspot.com/2011/09/notmnist-dataset.html}.

\bibitem[Byerly et~al.(2021)Byerly, Kalganova, and Dear]{byerly:mnist21}
Adam Byerly, Tatiana Kalganova, and Ian Dear.
\newblock No routing needed between capsules.
\newblock \emph{Neurocomputing}, 463:\penalty0 545--553, 2021.
\newblock ISSN 0925-2312.
\newblock \doi{https://doi.org/10.1016/j.neucom.2021.08.064}.
\newblock URL
  \url{https://www.sciencedirect.com/science/article/pii/S0925231221012546}.

\bibitem[Clanuwat et~al.(2018)Clanuwat, Bober-Irizar, Kitamoto, Lamb, Yamamoto,
  and Ha]{clanuwat:kmnist18}
Tarin Clanuwat, Mikel Bober-Irizar, Asanobu Kitamoto, Alex Lamb, Kazuaki
  Yamamoto, and David Ha.
\newblock Deep learning for classical {Japanese} literature.
\newblock In \emph{Neural Information Processing Systems 2018 Workshop on
  Machine Learning for Creativity and Design}, 2018.
\newblock URL \url{https://arxiv.org/abs/1812.01718}.

\bibitem[Cohen et~al.(2017)Cohen, Afshar, Tapson, and van
  Schaik]{cohen:emnist17}
Gregory Cohen, Saeed Afshar, Jonathan Tapson, and Andr{\'e} van Schaik.
\newblock {EMNIST: E}xtending {MNIST} to handwritten letters.
\newblock In \emph{2017 International Joint Conference on Neural Networks
  (IJCNN)}, pages 2921--2926, 2017.
\newblock \doi{10.1109/IJCNN.2017.7966217}.

\bibitem[Goodall and Swart(2022)]{goodall:zcta5-json}
John Goodall and Luke Swart.
\newblock {GeoJSON and TopoJSON} map files: zcta5.geo.json, 2022.
\newblock URL \url{https://github.com/jgoodall/us-maps/}.

\bibitem[Goodfellow et~al.(2014)Goodfellow, Bulatov, Ibarz, Arnoud, and
  Shet]{goodfellow:svhn14}
Ian~J. Goodfellow, Yaroslav Bulatov, Julian Ibarz, Sacha Arnoud, and Vinay
  Shet.
\newblock Multi-digit number recognition from street view imagery using deep
  convolutional neural networks.
\newblock In \emph{Proceedings of the International Conference on Learning
  Representations}, 2014.

\bibitem[Grother(2008)]{grother:nist-sd19}
Patrick~J. Grother.
\newblock {NIST Special Database} 19. {NIST} handprinted forms and characters
  database, 2008.
\newblock URL \url{https://www.nist.gov/srd/nist-special-database-19}.

\bibitem[LeCun et~al.(1998)LeCun, Bottou, Bengio, and Haffner]{lecun:mnist98}
Yann LeCun, L\'{e}on Bottou, Yoshua Bengio, and Patrick Haffner.
\newblock Gradient-based learning applied to document recognition.
\newblock \emph{Proceedings of the IEEE}, 86\penalty0 (11):\penalty0
  2278--2324, 1998.
\newblock \doi{10.1109/5.726791}.

\bibitem[Netzer et~al.(2011)Netzer, Wang, Coates, Bissacco, Wu, and
  Ng]{netzer:svhn11}
Yuval Netzer, Tao Wang, Adam Coates, Alessandro Bissacco, Bo~Wu, and Andrew~Y.
  Ng.
\newblock Reading digits in natural images with unsupervised feature learning.
\newblock In \emph{Proceedings of the NIPS Workshop on Deep Learning and
  Unsupervised Feature Learning}, 2011.

\bibitem[Newcomb(1881)]{newcomb:law1881}
Simon Newcomb.
\newblock Note on the frequency of use of the different digits in natural
  numbers.
\newblock \emph{American Journal of Mathematics}, 4\penalty0 (1):\penalty0
  39--40, 1881.

\bibitem[Nigrini(2012)]{nigrini:fraud12}
Mark~J. Nigrini.
\newblock \emph{{Benford's Law: A}pplications for Forensic Accounting,
  Auditing, and Fraud Detection}.
\newblock Wiley, first edition, April 2012.

\bibitem[Ross(2023)]{ross:randalyze23}
Jason Ross.
\newblock randalyze (v0.2.1) [software], 2023.
\newblock URL \url{https://pypi.org/project/randalyze/}.

\bibitem[{United States Postal Service}(2025)]{zipcodes25}
{United States Postal Service}.
\newblock {ZIP} codes by area and district codes, 2025.
\newblock URL \url{https://postalpro.usps.com/ZIP_Locale_Detail}.
\newblock Updated May 2, 2025; accessed on May 11, 2025.

\bibitem[Wagstaff(2012)]{wagstaff:matters12}
Kiri~L. Wagstaff.
\newblock Machine learning that matters.
\newblock In \emph{Proceedings of the 29th International Conference on Machine
  Learning}, pages 529--536, 2012.

\bibitem[Wilkinson et~al.(1992)Wilkinson, Geist, Janet, Grother, Burges,
  Creecy, Hammond, Hull, and Larsen]{wilkinson:nist92}
R.~Allen Wilkinson, Jon Geist, Stanley Janet, Patrick~J. Grother, Christopher
  J.~C. Burges, Robert Creecy, Bob Hammond, Jonathan~J. Hull, and Norman~L.
  Larsen.
\newblock The first census optical character recognition system conference.
\newblock NIST Interagency/Internal Report (NISTIR) 4912, National Institute of
  Standards and Technology, Gaithersburg, MD, 1992.

\bibitem[Xiao et~al.(2017)Xiao, Rasul, and Vollgraf]{xiao:fashion-mnist17}
Han Xiao, Kashif Rasul, and Roland Vollgraf.
\newblock {Fashion-MNIST: A} novel image dataset for benchmarking machine
  learning algorithms, 2017.
\newblock URL \url{https://arxiv.org/abs/1708.07747}.

\bibitem[Yadav and Bottou(2019)]{yadav:qmnist19}
Chhavi Yadav and L\'{e}on Bottou.
\newblock Cold case: {The} lost {MNIST} digits.
\newblock In \emph{Proceedings of the 33rd Conference on Neural Information
  Processing Systems (NeurIPS)}, 2019.

\end{thebibliography}


\end{document}